\documentclass[twoside,11pt,preprint]{article}

\usepackage{./myjmlr2e}

\usepackage[utf8]{inputenc}
\usepackage[T1]{fontenc}
\usepackage[english]{babel}
\usepackage[frozencache,cachedir=.]{minted} 
\usepackage{xspace}
\usepackage{graphicx}
\usepackage{paralist}
\usepackage[nolist]{acronym}
\usepackage{tabularx}
\usepackage{csquotes}
\usepackage{color}
\usepackage{colortbl}

\acrodef{ASR}{automatic speech recognition}
\acrodef{CI}{Continuous integration}
\acrodef{fMRI}{functional magnetic resonance imaging}
\acrodef{PCA}{principal component analysis}
\acrodef{PSD}{positive semidefinite}
\acrodef{SPD}{Symmetric positive definite}

\newcommand*{\topox}{\texttt{TopoX}\xspace}
\newcommand*{\toponetx}{\texttt{TopoNetX}\xspace}
\newcommand*{\topomodelx}{\texttt{TopoModelX}\xspace}
\newcommand*{\topoembedx}{\texttt{TopoEmbedX}\xspace}

\newcommand*{\Href}[2]{\href{#1}{\texttt{#2}}}

\definecolor{MidGray}{gray}{0.8}
\definecolor{LightGray}{gray}{0.9}

\usepackage{lastpage}
\jmlrheading{25}{2024}{1-\pageref{LastPage}}{1/24; Revised 11/24}{11/24}{24-0110}{\tiny{Mustafa Hajij,
Mathilde Papillon,
Florian Frantzen,
Jens Agerberg,
Ibrahem AlJabea,
Rub\'{e}n Ballester,
Claudio Battiloro, 
Guillermo Bern\'{a}rdez,
Tolga Birdal,
Aiden Brent,
Peter Chin,
Sergio Escalera, 
Simone Fiorellino,
Odin Hoff Gardaa,
Gurusankar Gopalakrishnan,
Devendra Govil,
Josef Hoppe,
Maneel Reddy Karri,
Jude Khouja,
Manuel Lecha,
Neal Livesay,
Jan Mei{\ss}ner,
Soham Mukherjee,
Alexander Nikitin,
Theodore Papamarkou,
Jaro Pr\'{i}lepok,
Karthikeyan Natesan Ramamurthy,
Paul Rosen,
Aldo Guzm\'{a}n-S\'{a}enz,
Alessandro Salatiello,
Shreyas N. Samaga,
Simone Scardapane,
Michael T. Schaub,
Luca Scofano,
Indro Spinelli,
Lev Telyatnikov,
Quang Truong,
Robin Walters,
Maosheng Yang,
Olga Zaghen,
Ghada Zamzmi,
Ali Zia,
Nina Miolane}}

\ShortHeadings{TopoX: A Suite of Python Packages for Machine Learning on Topological Domains}{Hajij et al.}

\begin{document}

\author{%
\name Mustafa Hajij\thanks{*\,Equal contribution as first author} \email mhajij@usfca.edu  \\
\name Mathilde Papillon\footnotemark[1] \email papillon@ucsb.edu \\
\name Florian Frantzen\footnotemark[1] \email florian.frantzen@cs.rwth-aachen.de \\
\name Jens Agerberg \email jensag@kth.se \\
\name Ibrahem AlJabea \email ialjab2@lsu.edu \\
\name Rub\'{e}n Ballester \email ruben.ballester@ub.edu \\
\name Claudio Battiloro \email cbattiloro@hsph.harvard.edu \\
\name Guillermo Bern\'{a}rdez \email guillermo.bernardez@upc.edu \\
\name Tolga Birdal \email tbirdal@imperial.ac.uk \\
\name Aiden Brent \email aidenjb81@gmail.com \\
\name Peter Chin \email peter.chin@dartmouth.edu \\
\name Sergio Escalera \email sescalera@ub.edu \\ 
\name Simone Fiorellino \email simone.fiorellino@uniroma1.it \\ 
\name Odin Hoff Gardaa \email odin.garda@uib.no \\
\name Gurusankar Gopalakrishnan \email ggopalakrishnan@dons.usfca.edu \\
\name Devendra Govil \email dgovil@dons.usfca.edu \\
\name Josef Hoppe \email hoppe@cs.rwth-aachen.de \\
\name Maneel Reddy Karri \email mkarri@dons.usfca.edu \\
\name Jude Khouja \email Jude@latynt.com\\
\name Manuel Lecha \email manuellecha@ub.edu \\
\name Neal Livesay \email nlivesay@ptc.com \\
\name Jan Mei{\ss}ner \email philipp.meissner@rwth-aachen.de \\
\name Soham Mukherjee \email mukher26@purdue.edu \\
\name Alexander Nikitin \email alexander.nikitin@aalto.fi \\
\name Theodore Papamarkou \email theo@zjnu.edu.cn \\
\name Jaro Pr\'{i}lepok \email jaroslav.prilepok@student.manchester.ac.uk \\
\name Karthikeyan Natesan Ramamurthy \email knatesa@us.ibm.com \\
\name Paul Rosen \email prosen@sci.utah.edu \\
\name Aldo Guzm\'{a}n-S\'{a}enz \email aldo.guzman.saenz@ibm.com \\ 
\name Alessandro Salatiello \email salatiello.alessandro@gmail.com \\
\name Shreyas N. Samaga \email ssamaga@purdue.edu \\
\name Simone Scardapane \email simone.scardapane@uniroma1.it \\
\name Michael T. Schaub \email schaub@cs.rwth-aachen.de \\ 
\name Luca Scofano \email luca.scofano@uniroma1.it \\
\name Indro Spinelli \email indro.spinelli@uniroma1.it \\
\name Lev Telyatnikov \email lev.telyatnikov@uniroma1.it \\
\name Quang Truong \email cong.minh.quang.truong.th@dartmouth.edu \\
\name Robin Walters \email r.walters@northeastern.edu \\
\name Maosheng Yang \email m.yang-2@tudelft.nl \\
\name Olga Zaghen \email o.zaghen@uva.nl \\
\name Ghada Zamzmi \email ghadh@mail.usf.edu \\
\name Ali Zia \email ali.zia@anu.edu.au \\
\name Nina Miolane \email ninamiolane@ucsb.edu
}

\editor{Joaquin Vanschoren}

\title{TopoX: A Suite of Python Packages for Machine Learning on Topological Domains \\\vspace{-2mm}}

\maketitle

\begin{abstract}
We introduce \topox, a Python software suite that provides reliable and user-friendly building blocks for computing and machine learning on topological domains that extend graphs: hypergraphs, simplicial, cellular, path and combinatorial complexes.
\topox consists of three packages: 
\texttt{\toponetx} facilitates constructing and computing on these domains, including working with nodes, edges and higher-order cells;
\texttt{\topoembedx} provides methods to embed topological domains into vector spaces, akin to popular graph-based embedding algorithms such as node2vec;
\texttt{\topomodelx} is built on top of \texttt{PyTorch} and offers a comprehensive toolbox of higher-order message passing functions for neural networks on topological domains. 
The extensively documented and unit-tested source code of \topox is available under MIT license at
  \textcolor{blue}{\Href{https://pyt-team.github.io/}{https://pyt-team.github.io/}}.
\end{abstract}


\section{Introduction}

Deep learning traditionally operates within Euclidean domains, focusing on structured data like images~\citep{krizhevsky2012imagenet} and sequences~\citep{sutskever2014sequence}. However, to handle more diverse data types, geometric deep learning~\citep[GDL;][]{bronstein2021geometric,zhou2020graph,cao2020comprehensive} has emerged. GDL extends deep learning to non-Euclidean data by leveraging geometric regularities like symmetries and invariances. Recently, topological deep learning~\citep[TDL;][]{hajijtopological} has gained attention, exploring models beyond traditional graph-based abstractions to process data with multi-way relations, such as simplicial complexes and hypergraphs. These extensions allow for the representation of diverse data domains encountered in scientific computations~\citep{feng2019hypergraph,schaub2020random,hajijcell,schaub2021signal,roddenberry2021principled,giusti2022cell,yang2023convolutional,barbarossa2020topological}. Despite theoretical advancements, practical implementation faces challenges due to the lack of accessible software libraries supporting deep learning models with higher-order structures.

\begin{figure}[!ht]
  \centering
  \includegraphics[width=\linewidth]{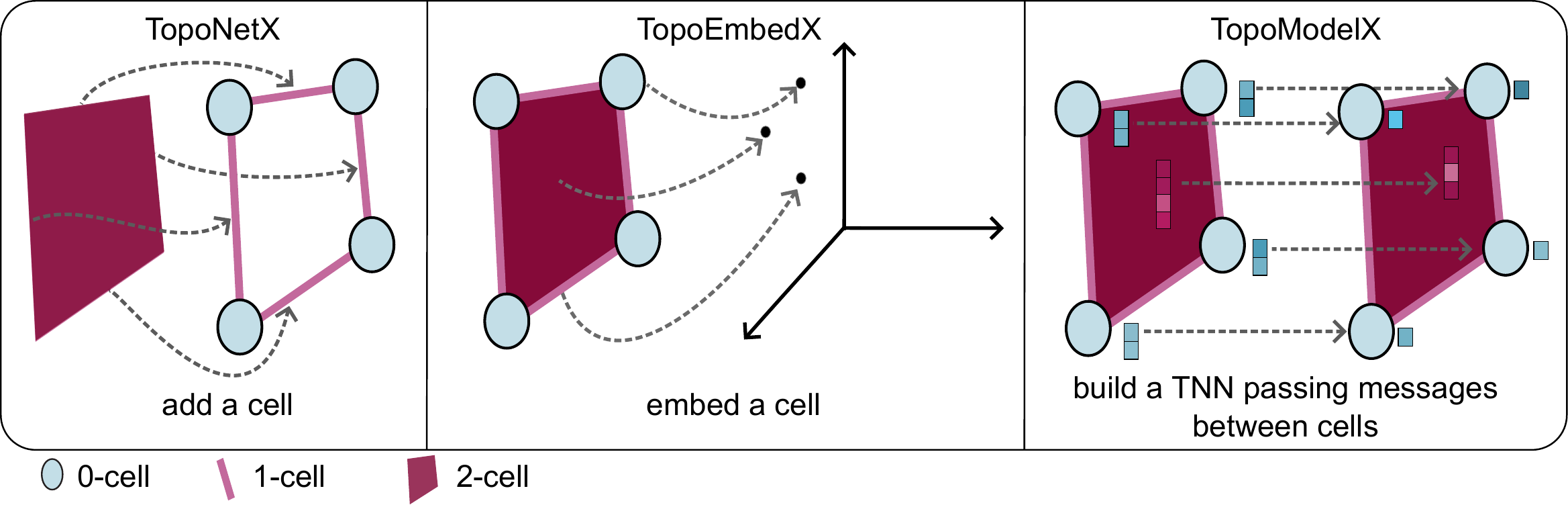}
  \caption{%
    Building blocks from the three packages of the \topox software suite. Left: \toponetx enables building topological domains such as cell complexes. The figure demonstrates adding a $2$-cell on a cell complex with \toponetx.
    Middle: \topoembedx enables embedding of topological domains inside a Euclidean space. The figure illustrates  embedding a $0$-cell, a $1$-cell and a $2$-cell from a cell complex inside a Euclidean space with \topoembedx. 
    Right: building a topological neural network (TNN) that processes data via higher-order message passing on a cell complex with \topomodelx.\vspace{-6mm}
  }
  \label{fig:visualization}
\end{figure}

In this paper, we present \topox, an open-source suite of Python packages designed for machine learning and deep learning operations in topological domains. 
\topox is organized into three Python packages: \toponetx, \topoembedx, and \topomodelx (see Figure~\ref{fig:visualization}). These packages enhance and generalize functionalities found in popular mainstream graph computations and learning tools, enabling them on topological domains. What sets \topox apart is its abstract general design and the exploitation of the resulting modeling flexibility in the implementation of a broad spectrum of topological domains and TDL models (see Table~\ref{tab:pkg_comp}). Further, every domain in \toponetx offers utilities to work with various components such as nodes, edges, and higher-order cells. \toponetx also supports computations using incidence matrices, (co)adjacency matrices, and up, down, and Hodge Laplacians. 
From a representation learning point of view, \topoembedx provides methods for embedding topological domains, or parts of these domains, into Euclidean domains. \topomodelx offers a wide range of TDL models based on a comprehensive implementation of higher-order message passing~\citep{hajijtopological}, built on the \texttt{PyTorch} framework~\citep{paszke2019}.  

The core objectives of \topox are to: facilitate topological research through foundational code and algorithm dissemination; broaden access to the field with user-friendly topological learning tools for the ML community; serve as a resource for topological learning enriched with examples, notebooks, and visualizations; and provide a unified API for topological domains. This API supports the generalization of topological spaces across scientific computations, enhancing interoperability, productivity, learning, and collaboration, while reducing maintenance, promoting code portability, and enabling parallel computing.



\section{Implementation Overview}\label{sec:geomstats}

The \toponetx package is organized into three main modules: \texttt{classes}, \texttt{algorithms}, and \texttt{transform}. The \texttt{classes} module implements numerous common topological domains, such as \texttt{SimplicialComplex}, \texttt{CellComplex}, and more; inheriting from the abstract class \texttt{Complex}. The \texttt{algorithms} module implements spectral methods, distance computation, and connected component analysis in topological domains. The \texttt{transform} module facilitates conversions between different topological domains. \toponetx uses \texttt{NumPy} and \texttt{SciPy} backends, and offers toy datasets and examples to facilitate learning and improve understanding.


The \topoembedx package supports the learning of representations for all topological domains available in \toponetx. This package contains the module \texttt{classes}, which implements topological representation learning algorithms that generalize popular graph-based representation learning algorithms, including \texttt{DeepCell} and \texttt{Cell2Vec}~\citep{hajijcell}. The \topoembedx package has an API inspired by 
\texttt{scikit-learn}~\citep{pedregosa2011scikit} and utilizes the KarateClub~\citep{karateclub} backend. While \topoembedx leverages methods from KarateClub, its functionalities are distinguished by a rigorous foundation in topological representation learning from~\cite{hajijtopological}. This foundation enables unique, theory-driven contributions, extending the package's capabilities to embed all  topological domains in \toponetx with a single unifying API, not just graphs as in KarateClub.



\topomodelx is a Python package for topological deep learning, providing efficient tools to implement topological neural networks (TNNs). The package consists of two main modules: \texttt{base} and \texttt{nn}. The \texttt{base} module implements higher-order message passing methods, allowing the construction of general-purpose TNNs using the tensor diagram formalism introduced in~\cite{hajijtopological} and surveyed in~\cite{mathilde2023}. The \texttt{nn} module implements various TNNs on popular topological domains, such as simplicial complexes, cell complexes, hypergraphs, and combinatorial complexes. Each implementation includes a corresponding Jupyter notebook tutorial for a user-friendly initiation into TDL. \topomodelx leverages \texttt{PyTorch} and \texttt{PyG} (PyTorch Geometric) backends~\citep{fey2019}. The \topomodelx package is the outcome of the unifying TDL framework introduced in~\cite{hajijtopological} which lead to the coding challenge that crowd-sourced the implementations of TDL models~\citep{papillon2023icml}.


\section{Comparison and Interaction with Other Packages}

The libraries most closely related to \toponetx are \texttt{NetworkX}~\citep{hagberg2008exploring} and \texttt{HypernetX}~\citep{liu2021parallel}. They facilitate computations on graphs and hypergraphs, respectively.
\toponetx utilizes a similar API to these two libraries to facilitate rapid adoption of topological domains, such as simplicial complexes, cell complexes, and colored hypergraphs. The \texttt{XGI} package~\citep{Landry_XGI_2023} offers hypergraph functionalities similar to \texttt{ HypernetX} with additional support for simplicial complexes and directed hypergraphs.


 The closest package to \texttt{TopoEmbedX} is \texttt{KarateClub}~\citep{karateclub}, which is a Python package consisting of methods for unsupervised learning on graph-structured data. \topoembedx extends the functionality of \texttt{KarateClub} to topological domains supported in \toponetx. \texttt{PyG} and \texttt{DGL}~\citep{wang2019}, which are two popular geometric deep learning packages that support deep learning models on graphs, are closely related to \topomodelx. The \texttt{DHG} (Deep HyperGraph) package~\citep{feng2019hypergraph} supports deep learning models defined on hypergraphs. All three of our packages feature a continuous integration pipeline and are well-tested with code coverage of  $ \geq 95\%$ per package. This is comparable or better than many related graph-based learning libraries, such as \texttt{PyG}, \texttt{DGL}, \texttt{XGI}, \texttt{NetworkX}, and \texttt{HyperNetX}. Table~\ref{tab:pkg_comp} provides a comparison between \topox and other Python packages.






\begin{table}[t]
  \footnotesize
  \centering
  \begin{tabularx}{\linewidth}{|l|>{\raggedright\arraybackslash}X|>{\raggedright\arraybackslash}X|}
    \hline
    \rowcolor{MidGray} \multicolumn{3}{|c|}{Comparison between \toponetx and other Python packages} \\ \hline
    \rowcolor{LightGray} \multicolumn{1}{|c|}{Packages}
    & \multicolumn{1}{c|}{Domains}
    & \multicolumn{1}{c|}{Operations} \\ \hline

    \texttt{\toponetx}
    & Graphs, colored hypergraphs, simplicial complexes, path complexes, cell complexes, combinatorial complexes
    & \texttt{add\_cell}, \texttt{add\_node}, \texttt{add\_simplex}, \texttt{adjacency\_matrix}, \texttt{coadjacency\_matrix}, \texttt{incidence\_matrix}, \texttt{hodge\_laplacian\_matrix}    \\ \hline

    \texttt{NetworkX}
    & Graphs
    & \texttt{add\_node}, \texttt{add\_edge}, \texttt{adjacency\_matrix}, \texttt{incidence\_matrix}, \texttt{laplacian\_matrix} \\ \hline

    \texttt{HyperNetX}
    & Hypergraphs
    & \texttt{add\_node}, \texttt{add\_edge}, \texttt{adjacency\_matrix}, \texttt{incidence\_matrix} \\ \hline

    \texttt{XGI}
    & Simplicial complexes, hypergraphs, dihypergraph
    &  \texttt{add\_node}, \texttt{add\_edge}, \texttt{adjacency\_matrix}, \texttt{boundary\_matrix}, \texttt{hodge\_laplacian} \\ \hline

    \rowcolor{MidGray} \multicolumn{3}{|c|}{Comparison between \topoembedx and other Python packages} \\ \hline
    \rowcolor{LightGray} \multicolumn{1}{|c|}{Packages}
    & \multicolumn{1}{c|}{Domains}
    & \multicolumn{1}{c|}{Embedding algorithms} \\ \hline

    \texttt{\topoembedx}
    & Graphs, colored hypergraphs, simplicial complexes, path complexes, cell complexes, combinatorial complexes
    & Cell2Vec, DeepCell, CellDiff2Vec, HigherOrderLaplacianEigenMap, HOPE, HOGLEE,  \\  \hline

    \texttt{Karateclub}
    & Graphs
    & Node2Vec, Graph2Vec, Diff2Vec, GL2Vec, IGE, Role2Vec, GraRep \\ \hline

    \rowcolor{MidGray} \multicolumn{3}{|c|}{Comparison between \topomodelx and other Python packages} \\ \hline
    \rowcolor{LightGray} \multicolumn{1}{|c|}{Packages}
    & \multicolumn{1}{c|}{Domains}
    & \multicolumn{1}{c|}{Push-forward operators} \\ \hline

    \texttt{\topomodelx}
    & Graphs, (colored) hypergraphs, simplicial complexes, path complexes, cell complexes, combinatorial complexes
    & (Higher order) message passing, merge operator, split operator \\ \hline

    \texttt{DGL}
    & Graphs
    & Message passing \\ \hline

    \texttt{DHG}
    & Graphs, hypergraphs
    & Message passing, hypergraph message passing \\ \hline

    \texttt{PyG}
    & Graphs
    & Message passing \\ \hline
  \end{tabularx}
  \caption{\texttt{TopoX} provides a user-friendly and comprehensive suite for building blocks and computing on topological domains. The table shows a comparison between \topox and other Python packages.\vspace{-4mm}}
  \label{tab:pkg_comp}
\end{table}











\section{Usage: Elementary Examples}

\toponetx provides a user-friendly interface which allows to create a complex in two main steps: first, instantiate the complex; second, add cells to that complex, as shown in the first three lines of the code snippet below. Furthermore, processing data on a complex requires matrices that describe the (co)adjacency of the incidence relations among cells.
\begin{small}
\begin{minted}{python}
import toponetx as tnx
cell_complex = tnx.CellComplex()
cell_complex.add_cell([1, 2, 3, 4], rank=2)
cell_complex.add_cell([1, 2, 5], rank=2)
L2 = cell_complex.hodge_laplacian_matrix(2)
\end{minted}
\end{small}



\noindent The following code snippet shows how \topoembedx embeds edges of the Stanford bunny dataset using the Cell2Vec algorithm~\citep{hajijcell}:
\begin{small}
    \begin{minted}{python}
import topoembedx as tex
cell_complex = tnx.datasets.stanford_bunny("cell")
model = tex.Cell2Vec()
model.fit(cell_complex, nbhd_type="adj", nbhd_dim={"adj": 1})
embeddings = model.get_embedding()
\end{minted}
\end{small}

\noindent The following code snippet shows how to instantiate, and run the forward-pass of an example of simplicial neural network~\citep[SAN;][]{battiloro2023generalized} with \topomodelx:
\begin{small}
\begin{minted}{python}
from topomodelx.nn.simplicial.san import SAN
from topomodelx.utils.sparse import from_sparse
sc = tnx.datasets.karate_club("simplicial")
x = torch.tensor([*sc.get_simplex_attributes("edge_feat").values()])
L1_down = from_sparse(sc.down_laplacian_matrix(rank=1))
L1_up = from_sparse(sc.up_laplacian_matrix(rank=1))
san_model = SAN(in_channels=2, hidden_channels=16, n_layers=2)
san_model(x, L1_down, L1_up)
\end{minted}
\end{small}
\toponetx provides a high-level declarative interface. In \topoembedx, each embedding algorithm is compatible with all complexes available in \toponetx. In \topomodelx, TNNs are classified according to the topological domain upon which they are defined. In each package, the directories that contain examples and notebooks offer an abundance of code snippets to assist users in starting their journey with \topox.

\newpage
\acks

M.~H. acknowledges support from the National Science Foundation, award DMS-2134231. N.~M. and M.~P. acknowledge support from the National Science Foundation, Award DMS-2134241.
S.~E. and R.~B. acknowledge support from Spanish project PID2022-136436NB-I00 and ICREA under the ICREA Academia programme. R.~B. acknowledges support from an FPU contract FPU21/00968, and by the Departament de Recerca i Universitats de la Generalitat de Catalunya (2021 SGR 00697).
T.~B. acknowledges support from the Engineering and Physical Sciences Research Council [grant EP/X011364/1].
T.~K.~D. acknowledges support from the National Science Foundation, Award CCF 2049010.
N.~L. acknowledges support from the Roux Institute and the Harold Alfond Foundation.
T.~P. acknowledges support from Zhejiang Normal University (distinguished professorship start-up grant YS304124985).
R.~W. acknowledges support from the National Science Foundation, Award DMS-2134178. 
P.~R. acknowledges support from the National Science Foundation, Award IIS-2316496.
M.~T.~S. and F.F. acknowledge funding by the Ministry of Culture and Science (MKW) of the German State of North Rhine-Westphalia (NRW R\"uckkehrprogramm). M.T.S and J.~H. acknowledge funding from the European Union (ERC, HIGH-HOPeS, 101039827). Views and opinions expressed are however those of M.~T.~S. only and do not necessarily reflect those of the European Union or the European Research Council Executive Agency; neither the European Union nor the granting authority can be held responsible for them. The authors acknowledge the ICML Topological Deep Learning challenge that jump-started the development of \texttt{TopoModelX}.
\bibliography{main}

\end{document}